\documentclass[letterpaper,10pt,conference]{ieeeconf}
\IEEEoverridecommandlockouts
\usepackage[T1]{fontenc}
\usepackage{amsmath,amssymb}
\usepackage{graphicx}
\usepackage{booktabs,tabularx,multirow}
\usepackage{xcolor}
\usepackage{cite}
\usepackage[hidelinks]{hyperref}
\hypersetup{pdftitle={FIERCE: From Generalist Robot Policies to Fast Specialists via Progress--Failure Feedback},
  pdfauthor={Runjia Tan, Yuang Tu, Yujie Yan, Lan Yu, Xuesong Tian, Chen Lv}, pdfsubject={Robot learning}, pdfkeywords={robot learning, progress--failure evaluator}}
\usepackage{microtype}
\usepackage{placeins}
\usepackage{balance}
\newcommand{\method}{\textsc{FIERCE}}
\newcommand{\E}{\mathbb{E}}

\newcommand{\sg}{\operatorname{sg}}
\newcommand{\BCE}{\operatorname{BCE}}

\newcommand{\NR}{\textsc{NR}}
\newcommand{\figref}[1]{Fig.~\ref{#1}}
\newcommand{\tabref}[1]{Table~\ref{#1}}
\newcommand{\secref}[1]{Sec.~\ref{#1}}

\makeatletter
\@ifundefined{bstctlcite}{%
  \newcommand{\bstctlcite}[1]{\@bsphack\if@filesw
    \immediate\write\@auxout{\string\citation{#1}}\fi\@esphack}
}{}
\makeatother

\title{\LARGE \bf FIERCE: From Generalist Robot Policies to Fast Specialists via Progress--Failure Feedback}
\author{Runjia Tan$^{1}$, Yuang Tu$^{1}$, Yujie Yan$^{1}$, Lan Yu$^{2}$, Xuesong Tian$^{2}$, Chen Lv$^{1}$%
\thanks{$^{1}$Nanyang Technological University.}%
\thanks{$^{2}$Guangzhou Cloudbutterfly Technology Co., Ltd.}%
}

\begin{document}
\maketitle
\thispagestyle{empty}
\pagestyle{empty}
\bstctlcite{BSTcontrol}

\begin{abstract}
Generalist robot policies offer useful initialization, but refining compact
specialists through limited physical interaction requires informative
learning feedback. We present FIERCE, a generalist-initialized reinforcement
learning framework centered on a unified, task-adaptive progress--failure
evaluator. Its architecture shares an observation--language representation
between an observed-progress head and an action-conditioned latent
predictor whose past and current predictions feed a causal sequence head
for task-failure estimation.
Joint supervision from progress and preference labels, synchronized
commands and observations, and terminal outcomes trains the evaluator;
target-task rollouts support adaptation and calibration. Fixed evaluator
snapshots provide progress shaping and failure-risk penalties alongside
independently verified terminal rewards, while evaluator and policy updates
alternate as new experience is collected. Refinement requires neither
continued generalist action queries nor a dedicated target-task simulator
or manually annotated dense rewards. Only the compact specialist is
retained at deployment. The evaluation separates feedback quality,
policy-learning efficiency, and deployment cost across simulation and two
contact-rich real tasks. Code, model weights, and data-restoration tools are released at \url{https://github.com/ar-mine/FIERCE}.
\end{abstract}

\section{Introduction}
\label{sec:intro}

Generalist robot policies, including vision--language--action models and
action-producing world models, provide transferable behavioral priors for
manipulation~\cite{ghosh2024octo,black2025pi05,ye2026dreamzero}. For repeated
insertion, alignment, and placement, however, broad task coverage is only
part of the deployment objective. A policy must respond to local variations
while meeting task-specific latency and computational constraints;
action chunking does not remove the importance of inference
latency~\cite{black2025rtc}. Compact, independently executable specialists
provide a complementary deployment choice and can serve as individual
reusable skills~\cite{tan2024multimodal}. Generalist demonstrations can
initialize such policies, but the central question is how they continue
to improve from their own physical experience.

\begin{figure*}[!tp]
  \centering
  \includegraphics[width=\textwidth]{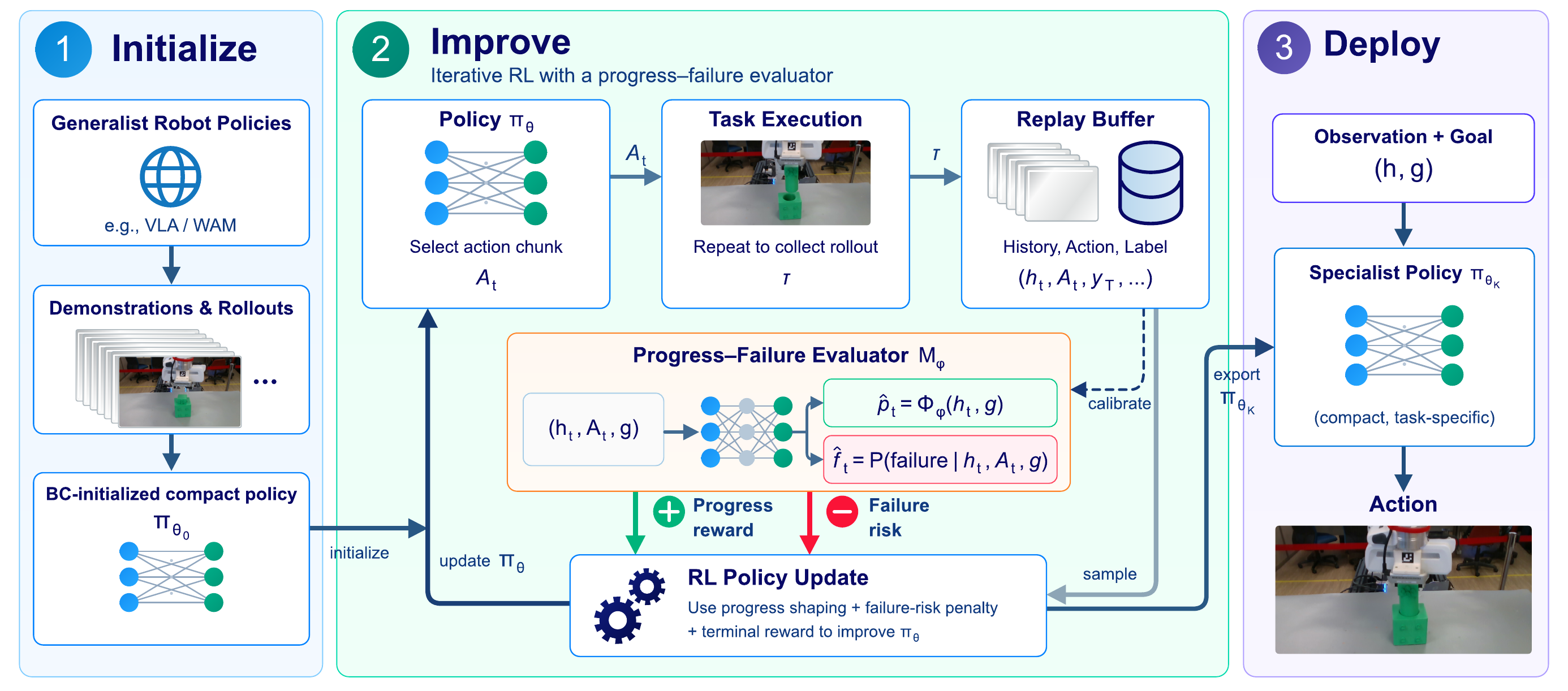}
  \caption{\textbf{FIERCE learning framework.}
  \textbf{(1)} Generalist demonstrations and rollouts initialize
  $\pi_{\theta_0}$. \textbf{(2)} Real trajectories and independently verified
  outcomes populate replay; the evaluator supplies progress shaping and
  action-conditioned failure penalties alongside terminal rewards.
  The dashed link denotes calibration, and scoring snapshots remain fixed
  during policy updates. Progress uses observed history; risk also depends
  on the action prefix, remaining time, and reference continuation in
  \eqref{eq:risk}. \textbf{(3)} Only $\pi_{\theta_K}$ is deployed; the
  generalist and evaluator are removed while independent safeguards remain.}
  \label{fig:overview}
\end{figure*}

Prior work combines imitation and reinforcement learning (RL) to acquire
compact experts from generalists~\cite{julg2025rpd}, while HIL-SERL
demonstrates policy improvement through real-world
interaction~\cite{luo2024hilserl}. Our focus is therefore the learning
feedback that enables refinement under a limited real-interaction budget,
rather than reproducing the generalist more faithfully. Independently
verified terminal outcomes establish whether an attempt succeeds, but do
not directly annotate which intermediate decisions should change. RL can
learn from these outcomes; the question is whether transferable progress
and failure predictions can provide more effective feedback when
only limited target-task experience is available.

The difficulty is that local progress and eventual success need not
agree. For example, pushing a misaligned connector toward a socket may
appear to advance insertion while increasing the chance of jamming.
Retracting to realign may temporarily reduce progress yet improve the
chance of completing the task. Progress feedback describes how much an
observed execution advances the task; prospective failure feedback
estimates the risk associated with a proposed action and subsequent
execution. A useful learning signal must distinguish these roles while
remaining grounded in real transitions and independently verified
outcomes.

Robometer demonstrates that progress and trajectory-preference
feedback can transfer across robot experience~\cite{liang2026robometer},
while Foresight shows that action-conditioned predictive latents can
support terminally supervised failure detection~\cite{zhang2026foresight}.
Motivated by these findings, we study a unified, task-adaptive
progress--failure evaluator as a feedback interface for specialist
learning. The central empirical question is whether learned failure
scores add value to progress shaping and terminal rewards during
physical refinement. Comparisons with progress-based and terminal-reward
learning assess the overall pipeline; a within-architecture ablation
isolates the use of the failure penalty while retaining failure-head
supervision.

We introduce \method{} (\emph{Failure-Informed Efficient Reinforcement
Learning for Compact Experts}), a generalist-initialized RL framework
for refining compact robot specialists through real-world interaction
(\figref{fig:overview}). Initialization supplies a compact
policy; subsequent improvement uses its own rollouts without requiring
further generalist action queries. The framework centers on a task-adaptive,
action-conditioned progress--failure evaluator. Its shared encoder
feeds an observed-progress head and a latent-prediction pathway with a
causal risk readout. Multi-task post-training uses diverse trajectories
and audited action-synchronized data; target-task fitting and score
calibration refresh this feedback using local successes and failures.
Progress increments, a failure-risk penalty, and verified
terminal rewards relabel real transitions for RL policy updates.
Evaluator and policy updates alternate as new experience becomes available.
Target-task learning requires no dedicated simulator, manually annotated
dense rewards, or separate sim-to-real policy-transfer stage. At
deployment, both the generalist and the large evaluator are removed,
leaving a standalone specialist.

Our contributions are twofold:
\begin{itemize}
    \item A unified progress--failure evaluator design with separate
    observed-progress and action-conditioned predictive pathways, trained
    as a refreshable feedback interface for specialist RL.

    \item A physical-refinement framework for compact policies initialized
    from demonstrations, including generalist rollouts, with no continued
    generalist action queries or dedicated target-task simulator.
    Deployment retains only the policy; within-architecture comparisons
    assess failure feedback and evaluator-training stages.
\end{itemize}
We assess progress and failure discrimination, feedback-guided policy
learning, and the resulting specialists' task-performance and inference-cost
trade-offs. Alternative-action validity is distinct from these evaluations.

\section{Related Work}
\label{sec:related}

\subsection{Robot Policy Adaptation and Specialization}

Robot policy improvement can retain the generalist or learn a separate
specialist. In the former route, RECAP improves a VLA using execution
experience and expert corrections~\cite{pi2025recap}, while RL Token
trains a lightweight actor--critic on pretrained VLA representations,
retaining the generalist during execution~\cite{xu2026rlt}.
Generalists themselves also remain strong alternatives: $\pi_{0.7}$
reports performance comparable to specialized RL policies on selected
tasks~\cite{pi2026pi07}. Specialization therefore targets a
performance--computation trade-off, rather than assuming that a
smaller policy is inherently more capable.

A complementary route uses generalist behavior to support a separate
expert. RPD combines behavioral cloning with teacher-action guidance
during RL exploration to refine compact task-specific
policies~\cite{julg2025rpd}, establishing that specialization can
extend beyond imitation. Real-world refinement is also supported by
HIL-SERL, which combines demonstrations, human corrections, and
sample-efficient RL for precise manipulation~\cite{luo2024hilserl}.
\method{} uses generalist trajectories for initialization but improves
an independently executable specialist without continued generalist
action queries. Our focus is the learning feedback needed for this
refinement under a limited real-interaction budget.

\begin{figure*}[!tp]
  \centering
  \includegraphics[width=\textwidth]{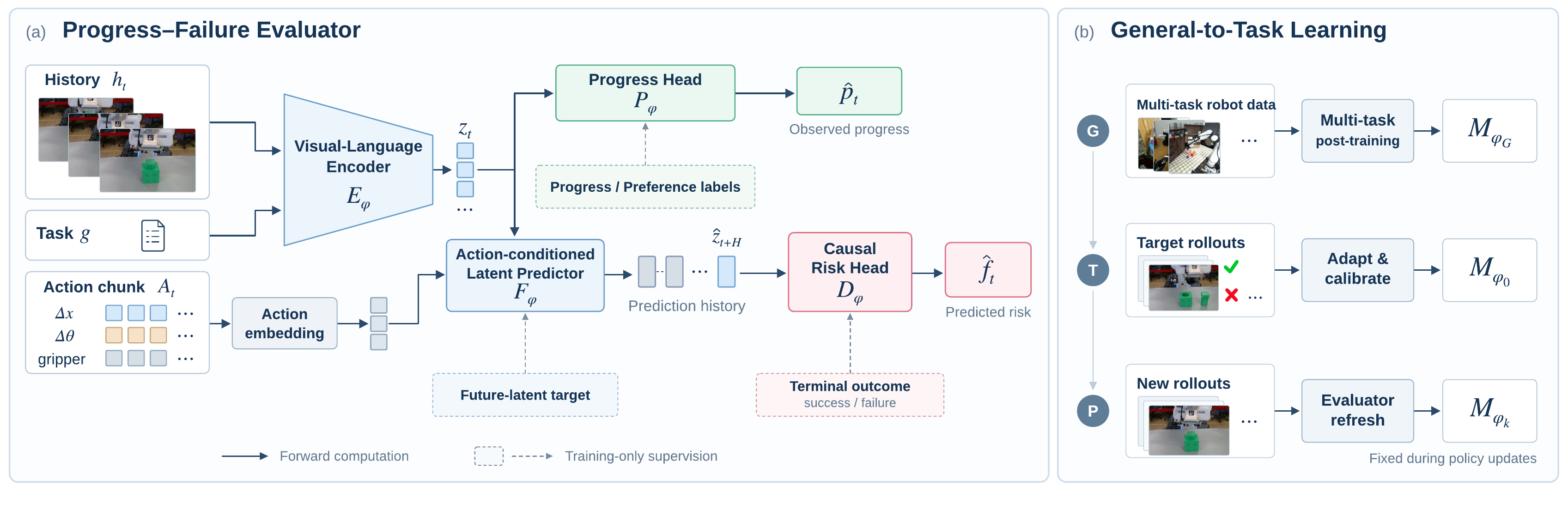}
  \caption{\textbf{Progress--failure evaluator and general-to-task learning.}
  \textbf{(a)} Shared history features feed a progress head and an
  action-conditioned predictor; its prediction history feeds the causal
  risk head. Dashed links denote training-only supervision, not future
  observations available at inference. \textbf{(b)} Multi-task post-training
  uses RBM progress/preference examples, audited action-synchronized data,
  and additional rollouts. Task adaptation and subsequent refreshes inherit
  the preceding parameters; fixed snapshots provide feedback for policy
  learning under the conditions in \eqref{eq:risk}.}
  \label{fig:critic}
\end{figure*}

\subsection{Reward and Failure Modeling for Policy Improvement}

Learned rewards provide a route from robot experience to transferable
supervision. ReWiND and RoboReward study language-conditioned rewards
for policy adaptation and evaluation~\cite{zhang2025rewind,lee2026roboreward},
while VLAC couples progress and completion feedback with real-world
RL~\cite{zhai2025vlac}. Closely related to our progress branch,
Robometer combines intra-trajectory progress and inter-trajectory
preferences across successful, suboptimal, and failed
experience~\cite{liang2026robometer}. Its reward model evaluates
observed execution without directly conditioning on a proposed action;
using failed trajectories for reward learning is thus distinct from
predicting candidate-action-conditioned failure risk.

Predictive representations extend supervision to future value and
action consequences. V-JEPA~2 learns action-conditioned latent dynamics
for robot planning from video and interaction data~\cite{assran2025vjepa2}.
ViVa uses video-generative representations for
value estimation~\cite{lv2026viva}. BORA explicitly conditions its
critic on VLM features and action chunks for offline value guidance,
then refines a frozen VLA through online residual
adaptation~\cite{chen2026bora}. These methods establish that
forward-looking or action-conditioned guidance is not unique to
failure prediction.

Failure models address a related but different objective: identifying
executions likely to fail. FAIL-Detect learns uncertainty-based signals
from successful demonstrations~\cite{xu2025faildetect}, while Gauge uses
video-world-model features for failure and anomaly
classification~\cite{ho2026gauge}. SAFE uses VLA internal features for
multitask detection~\cite{gu2025safe}, whereas Foresight combines
action-conditioned predictive latents with a causal detector trained
from terminal success/failure labels~\cite{zhang2026foresight}.
We instead use such predictions as training penalties. Whether a
failure score ranks alternative actions or remains calibrated after a
policy update is a separate empirical question. Risk-aware refinement is itself an established
setting: FARL couples a world-model-based safety critic with a
recovery policy to reduce intervention-requiring failures during
online learning~\cite{li2026farl}.

\method{} jointly trains observed-progress and action-conditioned
failure pathways as a training-time evaluator, with local adaptation
and score-calibration steps. It is not a runtime alarm or recovery
controller. The experiments assess progress/failure feedback within
this architecture and its use in compact-policy learning.

\section{FIERCE}
\label{sec:fierce}
\label{sec:method}

\subsection{Setting and Feedback Semantics}
\label{sec:problem}
For manipulation task $g$, let
$h_t=(o_{\leq t},q_{\leq t},a_{<t})$ contain visual observations,
proprioception, and executed commands. The decision context is
$x_t=(h_t,g,b_t)$, where $b_t=T_{\max}-t$ is the remaining task time.
Modules use declared observation windows and retain time information.
The command prefix $A_t=(a_{t|t},\ldots,a_{t+H-1|t})$ is selected before
execution for the next replan interval; its realized length $\ell\leq H$
is logged. Independently verified terminal outcomes
$y_\tau\in\{0,1\}$ indicate success; unknown outcomes are masked.
Action demonstrations $\mathcal D_0$ initialize the compact specialist
$\pi_\theta(A\mid x)$; in the generalist-initialized setting, these are
rollouts of $\pi_G$. The training-time progress--failure
evaluator $M_\phi$ supplies feedback for reward construction; a separate
RL critic $Q_\omega$ learns action values from the resulting rewards.
Refinement uses interaction budget $B_{\mathrm{int}}$, with units
specified per experiment.

Observed progress $\hat p_t$ estimates task advancement from history,
not success probability or unexecuted candidate actions. For collection
round $k$, risk is defined relative to continuation policy
$\pi_k=\pi_{\theta_k}$:
\begin{equation}
 f_k(h_t,A_t,g,b_t)
 =\Pr(y_\tau=0\mid h_t,A_t,g,b_t;\pi_k).
 \label{eq:risk}
\end{equation}
The estimate $\hat f_t$ concerns eventual failure after executing $A_t$
and continuing with $\pi_k$ within the remaining task time, not only the
next $H$ steps; it is not $1-\hat p_t$. Equation~\eqref{eq:risk} defines the estimation target. In the reported
learning experiments, $\hat f_t$ is used as a risk score, not as an
independently verified calibrated probability or causal action effect.
\method{} alternates evaluator adaptation and specialist improvement on
real experience, exporting only the compact policy
(\figref{fig:overview}).

\subsection{Progress--Failure Evaluator Architecture}
\label{sec:critic}
Our evaluator requires a shared representation for both task-progress estimation and action-conditioned predictive learning. We instantiate this representation with Cosmos~3 Edge, motivated by its development for physical-world understanding and prediction~\cite{nvidia2026cosmos3}. In our implementation, action conditioning is learned through an external latent predictor, while the progress branch follows
Robometer's progress and preference supervision~\cite{liang2026robometer}. Thus, we reuse the supervision principle rather than transferring Robometer's Qwen-specific prediction heads.

We instantiate the encoder with Cosmos~3 Edge observation--language
representations~\cite{nvidia2026cosmos3} and attach an external
action-conditioned latent predictor (\figref{fig:critic}). Robometer
provides a supervision formulation, not transferred backbone-specific
head weights. With $z_t=E_\phi(h_t,g)$,
\begin{align}
 \hat p_t &= P_\phi(z_t), &
 \hat z_{t+H} &= F_\phi(z_t,\operatorname{Enc}_a(A_t)), \nonumber\\
 \hat f_t &= D_\phi(\hat z_{\leq t+H},b_t).
 \label{eq:critic}
\end{align}
The progress branch evaluates observed state; the risk branch evaluates
predicted action consequences. Prediction history contains only
predictions formed at decisions no later than $t$, not subsequently
observed frames. The reference $\pi_k$ defines the continuation distribution,
not a policy-weight input. Future observations and outcomes provide training supervision only:
no future RGB is rendered at inference and ground-truth future frames
are never prediction inputs.

\subsection{General-to-Task Evaluator Learning}
\label{sec:data}
\textbf{Data and action restoration.}
RBM-1M is a multi-source robot-trajectory corpus for progress and
preference learning~\cite{liang2026robometer}. Its processed video--language
examples omit robot actions. We recover logged
commands from audited source episodes, including heterogeneous
Open X-Embodiment transitions~\cite{oxe2023}, and supplement them with
branching/policy rollouts. Restoration aligns parent timestamps, command
frames, units, controllers, and gripper conventions. Measured displacement
is not a command, and later feedback-dependent actions cannot form a chunk
known at $t$. Unknown action semantics mask action-dependent losses.
Rewound or instruction-mismatched videos are preference examples, not
verified physical failures. Views, subskills, and windows from one parent
episode share a split.

\textbf{General training.}
Available fields enable the masked objective
\begin{equation}
 \mathcal L_G=\lambda_p\mathcal L_p+
 \lambda_v\mathcal L_{\rm pref}+\lambda_d\mathcal L_{\rm dyn}
 +\lambda_f\mathcal L_f+\lambda_a\mathcal L_{\rm rank}.
 \label{eq:general_loss}
\end{equation}
Progress/preference retain RBM supervision. Dynamics uses
$\mathcal L_{\rm dyn}=\|\hat z_{t+H}-\sg(E_{\rm target}(o_{t+H}))\|_1$
with a stop-gradient target encoder. This target requires a fully
executed prefix and its future observation. Failure labels separately
require that the logged execution and continuation match
\eqref{eq:risk}; masking the dynamics loss does not by itself establish
that correspondence. For valid queries $\mathcal T_\tau$, terminal labels
train prefix risk:
\begin{equation}
 \mathcal L_f=
 \E_{\tau}\left[\frac{1}{|\mathcal T_\tau|}
 \sum_{t\in\mathcal T_\tau}
 \BCE(\hat f_t,1-y_\tau)\right].
 \label{eq:failure_loss}
\end{equation}
Episode normalization limits length bias. These labels identify
neither failure onset nor causal actions. Same-state simulator branches, with matched continuation
policy and remaining time, supply repeated success estimates $\bar y_i$.
Clearly ordered pairs add
\begin{equation}
 \mathcal L_{\rm rank}=
 -\log\sigma\bigl(\ell_f(A^-)-\ell_f(A^+)\bigr),
 \quad \bar y_+>\bar y_- ,
 \label{eq:ranking}
\end{equation}
where $\ell_f$ is the failure logit. Unrelated states with similar progress
are not counterfactual pairs.

\textbf{Task adaptation.}
Low-rank adapters~\cite{hu2021lora} and heads fit local successful/failed
rollouts; disjoint data fit failure-logit temperature and
bias~\cite{guo2017calibration}. Insufficient-data settings retain the
general calibrator. This is a score-calibration procedure; persistence of probability
calibration across policy updates is not established by the reported tests.

\begin{table*}[!tp]
\centering
\small
\setlength{\tabcolsep}{4pt}
\caption{Progress and failure prediction.
\textbf{A:} Action-free OOD progress and trajectory-quality ordering
(VOC and Kendall's $\tau_a$).
\textbf{B:} VLABench failure discrimination over three held-out task splits;
50\%-prefix AUROC uses the last complete query block by the midpoint.
Higher is better; these metrics do not measure probability calibration.
$\dagger$: reported by Robometer~\cite{liang2026robometer}.
$\ddagger$: evaluation under this work's VLABench protocol, not scores
quoted from the original benchmarks. $\S$: archived scores not reproduced
in a CPU rerun; see \secref{sec:critic_exp}.}
\label{tab:critic}
\begin{tabular}{lcc|lccc}
\toprule
\multicolumn{3}{c|}{A. OOD progress / quality ordering} &
\multicolumn{4}{c}{B. Held-out-task failure prediction} \\
Model &
VOC $\uparrow$ &
$\tau_a$ $\uparrow$ &
Model &
\shortstack{Rollout AUROC\\$\uparrow$} &
\shortstack{Prefix AUROC\\@50\% $\uparrow$} &
BalAcc $\uparrow$ \\
\midrule
ReWiND$^{\dagger}$
& 0.51 & 0.01
& FAIL-Detect/logpZO$^{\ddagger}$
& 0.577 & 0.654 & 0.557 \\

RoboReward-4B
& 0.84 & 0.51
& RND (two-head)$^{\ddagger}$
& 0.763 & 0.652 & 0.616 \\

Robometer-4B
& 0.94 & 0.64
& SAFE-LSTM$^{\ddagger,\S}$
& 0.733 & 0.577 & 0.571 \\

Cosmos-Reward (V+L)
& 0.95 & 0.47
& Gauge$^{\ddagger}$
& 0.799 & 0.612 & 0.603 \\

Cosmos-Reward-Cont.
& 0.94 & 0.38
& Foresight-Transformer$^{\ddagger}$
& 0.807 & 0.645 & 0.637 \\

\method{}
& \textbf{0.96} & \textbf{0.66}
& \method{} (full)
& \textbf{0.822} & \textbf{0.693} & \textbf{0.648} \\
\bottomrule
\end{tabular}
\end{table*}

\subsection{Evaluator-Guided Specialist Refinement}
\label{sec:rl}
\label{sec:init}
The specialist has its own visual encoder and action head. Successful
generalist trajectories or demonstrations initialize it by BC; failures
inform the evaluator. Stored demonstrations may regularize later learning,
without further generalist action queries.

For a real transition of length $\ell$, let $d_t$ denote true task
termination and $\Phi_\phi(h_t,g)=\hat p_t$. A fixed evaluator snapshot gives
\begin{align}
 \tilde r_t={}&\eta\,d_t(2y_\tau-1)-\rho\ell\Delta t-\beta\hat f_t \nonumber\\
 &+\alpha\bigl[(1-d_t)\gamma^\ell\Phi_\phi(h_{t+\ell},g)
                          -\Phi_\phi(h_t,g)\bigr].
 \label{eq:reward}
\end{align}
The outcome term is zero at nonterminal transitions. The progress
term has discounted potential-difference form with zero terminal
potential~\cite{ng1999shaping}. Risk and time penalties intentionally
change the objective; no policy-invariance claim is made for the full
objective. Summed risks are a regularizer, not an episode failure
probability or safety guarantee. The coefficient $\beta$ controls the
strength of this penalty, including possible over-conservatism.

\textbf{Iterative refinement.}
Each round collects real rollouts, updates/calibrates $M_\phi$, freezes a
versioned snapshot, consistently relabels replay, and updates $\pi_\theta$.
Recent-policy outcomes refresh risk; older data preserve progress/dynamics
knowledge rather than becoming labels for a new continuation policy.
After policy changes, risk remains provisional feedback until refreshed.
A SAC-style learner~\cite{haarnoja2018sac} mixes demonstration and online
replay following RLPD~\cite{ball2023rlpd} (\secref{sec:setup}). $M_\phi$ supplies
progress and risk scores for reward construction; the RL critic
$Q_\omega$ learns return through Bellman updates with discount $\gamma^\ell$
and mask $d_t$. Evaluator scoring can run asynchronously outside the
control path.

Success, the declared deadline, or predefined failure terminates a task.
Task termination is distinguished from external time-limit
truncation~\cite{pardo2018timelimits}. External stops/interventions do not automatically imply failure or
termination; unknown outcomes are masked. Bootstrap only from valid
pre-reset context when continuation is defined; otherwise exclude the
Bellman transition. Independent hardware safeguards remain.

\section{Experiments}
\label{sec:experiments}
We evaluate feedback quality (\tabref{tab:critic}), specialist performance
and deployment efficiency (\tabref{tab:policy}), and component contributions
to learning (\tabref{tab:ablation}).

\begin{figure*}[!tp]
  \centering
  \includegraphics[width=\textwidth]{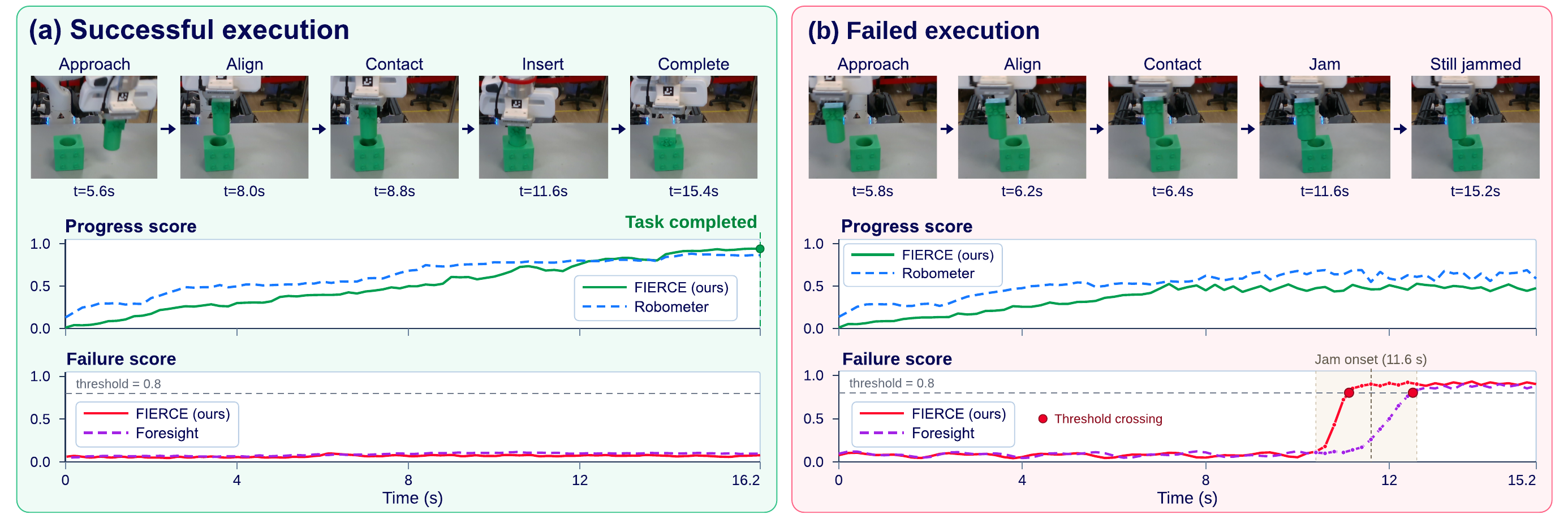}
  \caption{\textbf{Progress and failure feedback during peg-in-hole execution.}
  \textbf{(a)} Successful and \textbf{(b)} failed runs show selected frames
  with their timestamps, progress estimates (\method{}, Robometer), and
  failure scores (\method{}, Foresight). Later frames document outcomes,
  not prediction inputs. The 0.8 line is a displayed score reference;
  marked threshold crossings are offline diagnostics, not independently
  verified terminal failures or online recovery commands. Shared-threshold
  crossings alone do not establish matched false-alarm rates or calibrated
  detection lead times.}
  \label{fig:signals}
\end{figure*}

\subsection{Experimental Setup}
\label{sec:setup}
The evaluation spans LIBERO-90 Tasks 6, 28, 33, and 73~\cite{liu2023libero};
ManiSkill3 StackCube and PegInsertionSide~\cite{tao2025maniskill3}; and
real-world \emph{peg-in-hole} and \emph{cup stack} on Franka. Real specialists
learn through physical interaction; terminal success is verified
independently. Reward comparisons use the compact-policy setup below. The primary
within-architecture control retains evaluator supervision and sets the
failure-penalty coefficient to zero; external baselines compare complete
training pipelines rather than isolate a single architectural factor.
Action-free evaluator tests use only images and task text; action-aware tests
match command semantics and horizons. OOD refers to documented
evaluator-training exclusions, not verified exclusion from foundation pretraining.

\begin{figure*}[!tp]
  \centering
  \includegraphics[width=\textwidth]{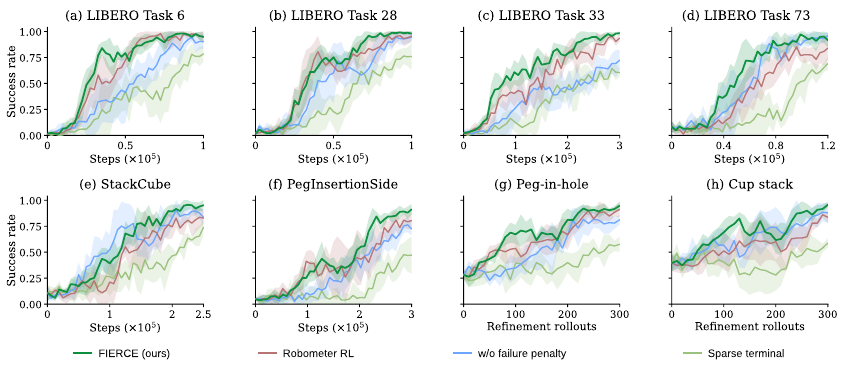}
  \caption{\textbf{Evaluator-guided policy learning.}
  \textbf{(a--d)} LIBERO Tasks 6, 28, 33, and 73;
  \textbf{(e,f)} ManiSkill3 StackCube and PegInsertionSide;
  \textbf{(g,h)} real peg-in-hole and cup stack.
  Success fractions are plotted against environment steps for simulation
  and refinement rollouts for the real tasks; horizons are task-specific.
  The four conditions are full \method{}, Robometer-guided RL, \method{}
  without its failure penalty ($\beta=0$, retaining failure supervision),
  and sparse terminal reward. Training uses seed~7; bands are not
  across-training-seed uncertainty estimates (see \secref{sec:setup}).}
  \label{fig:learning}
\end{figure*}

Simulation policies initialize from 50 successful demonstrations.
The horizons shown in \figref{fig:learning} are 100k steps for Tasks 6
and 28, 300k for Task 33, and 120k for Task 73. StackCube and
PegInsertionSide use displayed horizons of 250k and 300k steps. The BC-initialized RLPD-SAC actor uses a
frozen DINOv2-small encoder~\cite{oquab2023dinov2}, proprioception,
and three 512-unit hidden
layers with LayerNorm; the proposed horizon is four seven-dimensional
actions. Robometer is frozen, with isotonic calibration on a fixed
10-demonstration subset. All task-training conditions use seed~7, giving one training run per
task and condition. The two-task LIBERO protocol uses 200 final trials
per run (10 initial states, 20 repeats); this does not specify test
counts for the remaining tasks or measure variation across training
runs. The plotted bands have no documented aggregation unit and are
not used as inferential uncertainty. Results are descriptive single-run
comparisons; no training-seed significance claim is made.

The real-task protocol uses 300 refinement rollouts per
condition and evaluation every 30 rollouts. Tables~\ref{tab:policy}
and~\ref{tab:ablation} share final-success values where conditions overlap.
The 300-rollout figure specifies refinement, not the complete cost of
initialization, adaptation, calibration, evaluation, resets, or
interventions. It does not establish equal total target-task data use
or robot operating time across methods.

\subsection{Evaluator Generalization and Failure Prediction}
\label{sec:critic_exp}
\textbf{Protocol.}
Table~\ref{tab:critic}A compares action-free progress and quality ordering
using VOC on successful trajectories and Kendall's $\tau_a$. ReWiND values
are reproduced from Robometer~\cite{liang2026robometer}; other entries
come from archived evaluation artifacts. The \method{} checkpoint is
selected on development-set failure BCE subject to reward retention.
Table~\ref{tab:critic}B uses three fixed VLABench~\cite{zhang2024vlabench}
task splits held out from detector training. Baselines include an adapted
random-network-distillation detector~\cite{burda2018rnd}; its two-head
implementation is distinct from the original exploration objective. Rollout AUROC uses each execution's maximum score;
50\%-prefix AUROC uses the last complete query block by half of its actual
steps. Balanced accuracy averages conformal levels
$\alpha\in\{0.15,0.20,0.25\}$ within each split, then across splits.
This follows a Foresight-style protocol~\cite{zhang2026foresight}, not
an exact reproduction of its datasets or calibration selection.

\textbf{Prediction quality.}
\method{} reaches VOC/$\tau_a$ of 0.96/0.66 versus Robometer's 0.94/0.64.
Rollout AUROC is 0.822 versus 0.807 for Foresight-Transformer.
Prefix AUROC is 0.693 versus the strongest baseline value of 0.654
(FAIL-Detect/logpZO), a difference of 0.039; the difference from
Foresight-Transformer (0.645) is 0.048. Balanced accuracy is 0.648
versus 0.637 for Foresight-Transformer. These scores measure
outcome discrimination, not probability calibration or alternative-action
ranking. Archived scores and differing inference backends limit
cross-method attribution; the flagged SAFE-LSTM entry was not reproduced
by a CPU rerun. The gaps are not significance estimates.
\figref{fig:signals} illustrates the intended distinction between observed
progress and prospective risk.

\begin{table}[!tp]
\centering
\caption{Policy success and end-to-end latency.
\textbf{A:} Peg-in-hole/cup stack success rates and their equal-task mean.
\textbf{B:} Mean of task-level P95 observation-to-action latencies (ms)
across four LIBERO, two ManiSkill3, or two real tasks, measured on an
Intel Core i9-12900K CPU and NVIDIA RTX 5090 GPU.
$\dagger$ denotes the original/adapted model-pair mean, not an
individual model's latency.}
\label{tab:policy}
\begingroup
\small
\setlength{\tabcolsep}{3pt}
\renewcommand{\arraystretch}{1.10}
\begin{minipage}[t]{\columnwidth}
\centering
\textbf{A. Real-task performance}\par\smallskip
\begin{tabular*}{\linewidth}{@{\extracolsep{\fill}}lrrr@{}}
\toprule
\multirow{2}{*}{\textbf{Policy / training signal}}
& \multicolumn{3}{c}{\textbf{Success (\%) $\uparrow$}}\\
\cmidrule(lr){2-4}
& \shortstack{Peg-in-\\hole} & \shortstack{Cup\\stack} & Mean\\
\midrule
\multicolumn{4}{@{}l}{\textit{Generalist policies}}\\
Original $\pi_{0.5}$ & 40 & 50 & 45.0\\
Task-adapted $\pi_{0.5}$ & 78 & 82 & 80.0\\
\addlinespace[3pt]
\multicolumn{4}{@{}l}{\textit{Standalone compact policies}}\\
BC initialization only & 30 & 38 & 34.0\\
Robometer-guided RL & 80 & 86 & 83.0\\
\method{} w/o failure penalty & 82 & 88 & 85.0\\
\textbf{\method{} (full)} & \textbf{91} & \textbf{93} & \textbf{92.0}\\
\bottomrule
\end{tabular*}
\end{minipage}
\par\medskip
\begin{minipage}[t]{\columnwidth}
\centering
\textbf{B. Task-averaged end-to-end P95 (ms) $\downarrow$}\par\smallskip
\begin{tabular*}{\linewidth}{@{\extracolsep{\fill}}lrrr@{}}
\toprule
\textbf{Policy / timing condition} & LIBERO & ManiSkill3 & Real\\
\midrule
Original $\pi_{0.5}$ & 88.684 & 88.023 & \\
Original/adapted mean$^\dagger$ & & & 94.315\\
BC initialization only & 11.678 & 11.855 & 21.743\\
Robometer-guided RL & 11.715 & 11.232 & 21.343\\
\textbf{\method{} (full)} & 11.333 & 11.234 & 21.923\\
\bottomrule
\end{tabular*}
\end{minipage}
\endgroup
\end{table}

\subsection{Specialist Learning and Deployment}
\label{sec:learning_exp}
\label{sec:deploy_exp}

The four reward conditions are sparse terminal reward, Robometer-guided
RL, \method{} without its failure penalty, and full \method{}.
The full and no-penalty conditions share the compact actor, evaluator
supervision, terminal/time terms, and nominal refinement budget;
the latter sets only $\beta=0$. This tests using the learned risk penalty
within the same architecture, not the value of joint representation
learning. Robometer is frozen in the specified simulation setup, so that comparison
also changes evaluator adaptation and measures a pipeline-level difference.
Progress versus sparse rewards is a sanity check; it is not a new claim
that dense rewards can help RL.

\begin{figure}[!tbp]
  \centering
  \includegraphics[width=\columnwidth]{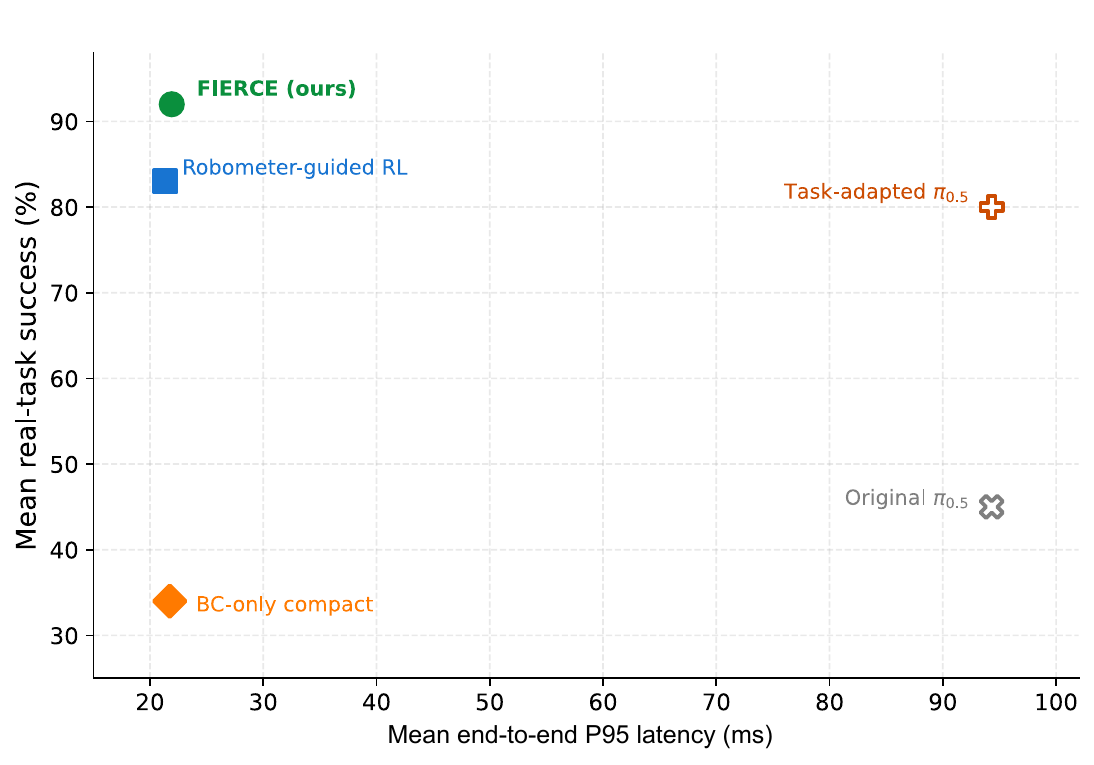}
  \caption{\textbf{Real-task success--latency comparison.}
  Success equally averages peg-in-hole and cup stack (\tabref{tab:policy}A);
  latency averages their end-to-end P95 values on the stated hardware
  (\tabref{tab:policy}B). Original and task-adapted $\pi_{0.5}$ share the
  reported 94.315\,ms model-pair mean on the horizontal axis; their
  individual latencies are not supplied. This plot does not measure
  task throughput.}
  \label{fig:deployment}
\end{figure}

\begin{table*}[!tp]
\centering
\caption{Component contributions to specialist learning.
Peg-in-hole and cup stack use the same compact actor architecture and a
300-rollout refinement budget per task.
\textbf{A:} Feedback-design ablations.
\textbf{B:} Evaluator-training ablations.
Higher final success and learning-curve AUC, and lower $N_{80}$, are preferred.}
\label{tab:ablation}
\begingroup
\small
\setlength{\tabcolsep}{4pt}
\renewcommand{\arraystretch}{1.10}
\begin{tabular*}{\textwidth}{@{\extracolsep{\fill}}lrrr!{\hspace{7pt}\vrule width 0.3pt\hspace{7pt}}rrr@{}}
\toprule
\textbf{Variant} & \multicolumn{3}{c}{\textbf{Peg-in-hole}}
& \multicolumn{3}{c}{\textbf{Cup stack}} \\
\cmidrule(lr){2-4}\cmidrule(l){5-7}
& \shortstack{Final success\\(\%) $\uparrow$}
& \shortstack{Norm. AUC\\$[0,1]$ $\uparrow$}
& \shortstack{$N_{80}$\\(rollouts) $\downarrow$}
& \shortstack{Final success\\(\%) $\uparrow$}
& \shortstack{Norm. AUC\\$[0,1]$ $\uparrow$}
& \shortstack{$N_{80}$\\(rollouts) $\downarrow$} \\
\midrule
\textbf{\method{} (full)} & \textbf{91} & \textbf{0.733} & \textbf{185} & \textbf{93} & \textbf{0.786} & \textbf{147}\\
\addlinespace[3pt]
\multicolumn{7}{@{}l}{\textit{A. Feedback design}} \\
\quad w/o failure penalty ($\beta=0$) & 82 & 0.634 & 270 & 88 & 0.705 & 206\\
\quad w/o candidate-action conditioning & 78 & 0.598 & \NR & 84 & 0.674 & NR\\
\quad w/o action-ranking loss ($\lambda_a=0$) & 88 & 0.706 & 186 & 91 & 0.762 & 152\\
\addlinespace[3pt]
\multicolumn{7}{@{}l}{\textit{B. Evaluator training procedure}} \\
\quad w/o general evaluator post-training (G) & 74 & 0.550 & \NR & 81 & 0.632 & NR\\
\quad w/o initial task adaptation (T) & 77 & 0.581 & \NR & 85 & 0.691 & NR\\
\quad Frozen task evaluator (no P refresh) & 83 & 0.684 & 213 & 89 & 0.754 & 154\\
\bottomrule
\end{tabular*}
\par\smallskip
\begin{minipage}{\textwidth}
\footnotesize
\textit{Notes.} AUC is trapezoidal area under the success-fraction
learning curve divided by 300, not detection AUROC. $N_{80}$ is the first
30-rollout evaluation checkpoint reaching 80\% success, without
interpolation; \NR{} means unreached within budget. Shared final successes
match \tabref{tab:policy}A. Ablation definitions are in the text.
\end{minipage}
\endgroup
\end{table*}

Table~\ref{tab:policy}A reports success on the two real tasks.
The task-adapted generalist uses successful-rollout SFT, not VLA RL.
In the reported seed-7 runs, full \method{} averages 92\% success,
versus 83\% for Robometer-guided RL and 85\% without the failure penalty.
Per-checkpoint test counts and the alignment of \figref{fig:learning}
with the table summaries are not fully documented; the reported
AUC and threshold costs are therefore descriptive, not independently
reconstructed from the displayed traces.

Table~\ref{tab:policy}B reports task-averaged end-to-end P95 latency:
four LIBERO tasks, two ManiSkill3 tasks, and peg-in-hole/cup stack.
Each entry averages per-task P95 values, rather than pooling queries
before taking a percentile. All timings use the i9-12900K/RTX 5090.
The 94.315\,ms Real entry is the original/adapted $\pi_{0.5}$
model-pair average; Fig.~\ref{fig:deployment} assigns that aggregate
coordinate to both markers, not individual model measurements.
FIERCE's 21.923\,ms is close to BC-only's 21.743\,ms and
Robometer-guided RL's 21.343\,ms. Low inference cost primarily follows
from the shared compact actor and removal of the evaluator at deployment;
it is not evidence that failure feedback accelerates inference, evaluator
training, or task throughput.

\subsection{Component Ablations}
\label{sec:ablation}
Table~\ref{tab:ablation} separates feedback design from evaluator training,
using the same actor and per-task budget. Final success is evaluated at
300 rollouts; normalized AUC is the trapezoidal area under the
success-fraction curve divided by 300, not a detection AUROC. $N_{80}$
is the first evaluated checkpoint reaching 80\% success, without
interpolation; \NR{} denotes censoring, not a cost of 300. Real-task $N_{80}$ counts rollout checkpoints and is not pooled with
simulation environment-step budgets.

\textbf{Definitions.}
Setting $\beta=0$ removes only the risk penalty, retaining failure-head
training and terminal reward. The no-action variant retrains the evaluator without candidate commands;
actor actions remain unchanged. This is an evaluator-configuration
ablation: retraining can change shared features and auxiliary objectives,
so its effect cannot be attributed solely to removing one risk-head input. Setting
$\lambda_a=0$ removes ranking supervision while retaining branch data for
other valid losses. Skipping G retains the foundation checkpoint and T/P;
skipping initial T retains the general calibrator, local replay, and later
P refreshes. The frozen variant completes T, then fixes $M_\phi$ and its
calibrator while the actor and RL critic $Q_\omega$ continue learning. These ablations compare configurations within the same nominal refinement
budget, not an equalized end-to-end data or computation budget.

For full versus $\beta=0$, final success is 91\%/93\% versus
82\%/88\%, and $N_{80}$ is 180/150 versus 270/210 on peg/cup.
These observations support using risk feedback in the reported runs.
The no-action configuration performs worse than $\beta=0$, but the
present experiments do not identify the cause of that difference.

\FloatBarrier
\section{Discussion and Conclusion}
\label{sec:conclusion}
\method{} uses a unified progress--failure evaluator as a training-time
interface for compact-policy refinement. Observed-progress estimation and
action-conditioned latent prediction provide different learning signals;
local adaptation and refreshes fit them to new execution data. The
reported within-architecture ablation favors using the failure penalty
on both real tasks, while deployment retains only the compact actor.
These are single-seed observations, not evidence of robustness to retraining
or reduced total robot operating cost.

The study does not isolate parameter sharing from independently trained
progress and failure models, and its discrimination and RL metrics do not
establish policy-wise probability calibration or same-state alternative-action
ranking. Risk penalties can discourage useful corrective actions; policy shift
and hidden contact states remain limitations. We claim no formal safety,
optimal-policy invariance, or early-recovery guarantee. The latency
measurement concerns the deployed actor, not evaluator compute or
successful task throughput. Data, weights, restoration tools, split manifests, and evaluation
scripts will be released subject to source licenses.

\begin{samepage}
\section*{Acknowledgments}
\textbf{Generative-AI assistance.} ChatGPT (OpenAI) supported drafting
and editing (Abstract, Sections I--V and captions); code development,
debugging, experiment planning and analysis scripts (Sections III--IV);
and initial figure/layout and illustrative-asset prototyping. The authors
retain responsibility for the code, measurements and conclusions.\par
\end{samepage}

\FloatBarrier
\balance
\bibliographystyle{IEEEtran}
\bibliography{references}

\begin{thebibliography}{10}
\providecommand{\url}[1]{#1}
\csname url@samestyle\endcsname
\providecommand{\newblock}{\relax}
\providecommand{\bibinfo}[2]{#2}
\providecommand{\BIBentrySTDinterwordspacing}{\spaceskip=0pt\relax}
\providecommand{\BIBentryALTinterwordstretchfactor}{4}
\providecommand{\BIBentryALTinterwordspacing}{\spaceskip=\fontdimen2\font plus
\BIBentryALTinterwordstretchfactor\fontdimen3\font minus
  \fontdimen4\font\relax}
\providecommand{\BIBforeignlanguage}[2]{{%
\expandafter\ifx\csname l@#1\endcsname\relax
\typeout{** WARNING: IEEEtran.bst: No hyphenation pattern has been}%
\typeout{** loaded for the language `#1'. Using the pattern for}%
\typeout{** the default language instead.}%
\else
\language=\csname l@#1\endcsname
\fi
#2}}
\providecommand{\BIBdecl}{\relax}
\BIBdecl

\bibitem{ghosh2024octo}
D.~Ghosh, H.~R. Walke, K.~Pertsch \emph{et~al.}, ``{Octo}: An open-source
  generalist robot policy,'' in \emph{Proceedings of Robotics: Science and
  Systems}, Delft, Netherlands, 2024.

\bibitem{black2025pi05}
{Physical Intelligence}, K.~Black, N.~Brown \emph{et~al.}, ``{$\pi_{0.5}$}: A
  vision-language-action model with open-world generalization,'' \emph{arXiv
  preprint arXiv:2504.16054}, 2025.

\bibitem{ye2026dreamzero}
S.~Ye, Y.~Ge, K.~Zheng \emph{et~al.}, ``World action models are zero-shot
  policies,'' \emph{arXiv preprint arXiv:2602.15922}, 2026.

\bibitem{black2025rtc}
K.~Black, M.~Y. Galliker, and S.~Levine, ``Real-time execution of action
  chunking flow policies,'' \emph{arXiv preprint arXiv:2506.07339}, 2025.

\bibitem{tan2024multimodal}
R.~Tan, S.~Lou, Y.~Zhou, and C.~Lv, ``Multi-modal {LLM}-enabled long-horizon
  skill learning for robotic manipulation,'' in \emph{2024 IEEE International
  Conference on Cybernetics and Intelligent Systems (CIS) and IEEE
  International Conference on Robotics, Automation and Mechatronics
  (RAM)}.\hskip 1em plus 0.5em minus 0.4em\relax IEEE, 2024, pp. 14--19.

\bibitem{julg2025rpd}
T.~J{\"u}lg, W.~Burgard, and F.~Walter, ``Refined policy distillation: From
  {VLA} generalists to {RL} experts,'' \emph{arXiv preprint arXiv:2503.05833},
  2025.

\bibitem{luo2024hilserl}
J.~Luo, C.~Xu, J.~Wu, and S.~Levine, ``Precise and dexterous robotic
  manipulation via human-in-the-loop reinforcement learning,'' \emph{arXiv
  preprint arXiv:2410.21845}, 2024.

\bibitem{liang2026robometer}
A.~Liang, Y.~Korkmaz, J.~Zhang \emph{et~al.}, ``{Robometer}: Scaling
  general-purpose robotic reward models via trajectory comparisons,''
  \emph{arXiv preprint arXiv:2603.02115}, 2026.

\bibitem{zhang2026foresight}
H.~Zhang, Y.~Lu, B.~Wang \emph{et~al.}, ``{Foresight}: Failure detection for
  long-horizon robotic manipulation with action-conditioned world model
  latents,'' \emph{arXiv preprint arXiv:2606.23085}, 2026.

\bibitem{pi2025recap}
{Physical Intelligence}, ``{$\pi^{*}_{0.6}$}: A {VLA} that learns from
  experience,'' \emph{arXiv preprint arXiv:2511.14759}, 2025.

\bibitem{xu2026rlt}
C.~Xu, J.~T. Springenberg, M.~Equi \emph{et~al.}, ``{RL Token}: Bootstrapping
  online {RL} with vision-language-action models,'' \emph{arXiv preprint
  arXiv:2604.23073}, 2026.

\bibitem{pi2026pi07}
{Physical Intelligence}, ``{$\pi_{0.7}$}: A steerable generalist robotic
  foundation model with emergent capabilities,'' \emph{arXiv preprint
  arXiv:2604.15483}, 2026.

\bibitem{zhang2025rewind}
J.~Zhang, Y.~Luo, A.~Anwar \emph{et~al.}, ``{ReWiND}: Language-guided rewards
  teach robot policies without new demonstrations,'' in \emph{Proceedings of
  the 9th Conference on Robot Learning}, ser. Proceedings of Machine Learning
  Research, vol. 305.\hskip 1em plus 0.5em minus 0.4em\relax PMLR, 2025, pp.
  460--488.

\bibitem{lee2026roboreward}
T.~Lee, A.~Wagenmaker, K.~Pertsch, P.~Liang, S.~Levine, and C.~Finn,
  ``{RoboReward}: General-purpose vision-language reward models for robotics,''
  \emph{arXiv preprint arXiv:2601.00675}, 2026.

\bibitem{zhai2025vlac}
S.~Zhai, Q.~Zhang, T.~Zhang \emph{et~al.}, ``A vision-language-action-critic
  model for robotic real-world reinforcement learning,'' \emph{arXiv preprint
  arXiv:2509.15937}, 2025.

\bibitem{assran2025vjepa2}
M.~Assran, A.~Bardes, D.~Fan \emph{et~al.}, ``{V-JEPA 2}: Self-supervised video
  models enable understanding, prediction and planning,'' \emph{arXiv preprint
  arXiv:2506.09985}, 2025.

\bibitem{lv2026viva}
J.~Lv, H.~Li, J.~Li \emph{et~al.}, ``{ViVa}: A video-generative value model for
  robot reinforcement learning,'' \emph{arXiv preprint arXiv:2604.08168}, 2026.

\bibitem{chen2026bora}
Z.~Chen, Y.~Han, Y.~Shao \emph{et~al.}, ``{BORA}: Bridging offline
  reinforcement learning and online residual adaptation for real-world
  dexterous {VLA} models,'' \emph{arXiv preprint arXiv:2605.30226}, 2026.

\bibitem{xu2025faildetect}
C.~Xu, T.~K. Nguyen, E.~Dixon \emph{et~al.}, ``Can we detect failures without
  failure data? uncertainty-aware runtime failure detection for imitation
  learning policies,'' in \emph{Proceedings of Robotics: Science and Systems},
  2025.

\bibitem{ho2026gauge}
M.~Ho, M.~F. Ginting, I.~R. Ward \emph{et~al.}, ``World model failure
  classification and anomaly detection for autonomous inspection,'' \emph{arXiv
  preprint arXiv:2602.16182}, 2026.

\bibitem{gu2025safe}
Q.~Gu, Y.~Ju, S.~Sun \emph{et~al.}, ``{SAFE}: Multitask failure detection for
  vision-language-action models,'' in \emph{Advances in Neural Information
  Processing Systems}, vol.~38, 2025.

\bibitem{li2026farl}
H.~Li, K.~Lei, S.~Zang \emph{et~al.}, ``Failure-aware {RL}: Reliable
  offline-to-online reinforcement learning with self-recovery for real-world
  manipulation,'' \emph{arXiv preprint arXiv:2601.07821}, 2026.

\bibitem{nvidia2026cosmos3}
{NVIDIA}, ``{Cosmos 3}: Omnimodal world models for physical {AI},'' \emph{arXiv
  preprint arXiv:2606.02800}, 2026.

\bibitem{oxe2023}
{Open X-Embodiment Collaboration}, ``Open {X}-embodiment: Robotic learning
  datasets and {RT-X} models,'' \emph{arXiv preprint arXiv:2310.08864}, 2023.

\bibitem{hu2021lora}
E.~J. Hu, Y.~Shen, P.~Wallis \emph{et~al.}, ``{LoRA}: Low-rank adaptation of
  large language models,'' \emph{arXiv preprint arXiv:2106.09685}, 2021.

\bibitem{guo2017calibration}
C.~Guo, G.~Pleiss, Y.~Sun, and K.~Q. Weinberger, ``On calibration of modern
  neural networks,'' in \emph{Proceedings of the 34th International Conference
  on Machine Learning}, ser. Proceedings of Machine Learning Research,
  vol.~70.\hskip 1em plus 0.5em minus 0.4em\relax PMLR, 2017, pp. 1321--1330.

\bibitem{ng1999shaping}
A.~Y. Ng, D.~Harada, and S.~J. Russell, ``Policy invariance under reward
  transformations: Theory and application to reward shaping,'' in
  \emph{Proceedings of the Sixteenth International Conference on Machine
  Learning}.\hskip 1em plus 0.5em minus 0.4em\relax Morgan Kaufmann, 1999, pp.
  278--287.

\bibitem{haarnoja2018sac}
T.~Haarnoja, A.~Zhou, P.~Abbeel, and S.~Levine, ``Soft actor-critic: Off-policy
  maximum entropy deep reinforcement learning with a stochastic actor,'' in
  \emph{Proceedings of the 35th International Conference on Machine Learning},
  ser. Proceedings of Machine Learning Research, vol.~80.\hskip 1em plus 0.5em
  minus 0.4em\relax PMLR, 2018, pp. 1861--1870.

\bibitem{ball2023rlpd}
P.~J. Ball, L.~Smith, I.~Kostrikov, and S.~Levine, ``Efficient online
  reinforcement learning with offline data,'' in \emph{Proceedings of the 40th
  International Conference on Machine Learning}, ser. Proceedings of Machine
  Learning Research, vol. 202, 2023, pp. 1577--1594.

\bibitem{pardo2018timelimits}
F.~Pardo, A.~Tavakoli, V.~Levdik, and P.~Kormushev, ``Time limits in
  reinforcement learning,'' in \emph{Proceedings of the 35th International
  Conference on Machine Learning}, ser. Proceedings of Machine Learning
  Research, vol.~80, 2018, pp. 4045--4054.

\bibitem{liu2023libero}
B.~Liu, Y.~Zhu, C.~Gao \emph{et~al.}, ``{LIBERO}: Benchmarking knowledge
  transfer for lifelong robot learning,'' \emph{arXiv preprint
  arXiv:2306.03310}, 2023.

\bibitem{tao2025maniskill3}
S.~Tao, F.~Xiang, A.~Shukla \emph{et~al.}, ``{ManiSkill3}: {GPU} parallelized
  robotics simulation and rendering for generalizable embodied {AI},'' in
  \emph{Robotics: Science and Systems}, 2025.

\bibitem{oquab2023dinov2}
M.~Oquab, T.~Darcet, T.~Moutakanni \emph{et~al.}, ``{DINOv2}: Learning robust
  visual features without supervision,'' \emph{arXiv preprint
  arXiv:2304.07193}, 2023.

\bibitem{zhang2024vlabench}
S.~Zhang, Z.~Xu, P.~Liu \emph{et~al.}, ``{VLABench}: A large-scale benchmark
  for language-conditioned robotics manipulation with long-horizon reasoning
  tasks,'' \emph{arXiv preprint arXiv:2412.18194}, 2024.

\bibitem{burda2018rnd}
Y.~Burda, H.~Edwards, A.~Storkey, and O.~Klimov, ``Exploration by random
  network distillation,'' \emph{arXiv preprint arXiv:1810.12894}, 2018.

\end{thebibliography}
\end{document}